\documentclass{article} 
\usepackage{spconf}
\usepackage{latexsym}
\usepackage{amssymb}
\usepackage{cite}

\usepackage{fancyhdr, ifpdf}
\usepackage[dvipdfm]{graphicx}
\usepackage{epstopdf}
\usepackage{amsmath}
\usepackage{bm}
\usepackage{float}
\usepackage{inouelab}
\usepackage{tabularx}

\newcommand{\BernoulliDist}{\mathop{\mathrm{Bernoulli}}\limits}
\newcommand{\GammaDist}{\mathop{\mathrm{Gamma}}\limits}
\newcommand{\NormalDist}{\mathop{\calN}\limits}
\newcommand{\stepOO}{{(\!0\!)}}
\newcommand{\steptt}{{(\!t\!)}}
\newcommand{\steptl}{{(\!t\!+\!1\!)}}

\title{POSTERIOR MEAN SUPER-RESOLUTION\\
WITH A COMPOUND GAUSSIAN MARKOV RANDOM FIELD PRIOR}

\name
  {Takayuki Katsuki, Masato Inoue}
	\address{Department of Electrical Engineering and Bioscience, Waseda University\\
3--4--1, Okubo, Shinjuku, Tokyo 1698555, Japan \\ 
}
\begin{document}
\ninept
\maketitle

\begin{abstract}
This manuscript proposes a posterior mean (PM) super-resolution (SR) method with a compound Gaussian Markov random field (MRF) prior. SR is a technique to estimate a spatially high-resolution image from observed multiple low-resolution images. A compound Gaussian MRF model provides a preferable prior for natural images that preserves edges. PM is the optimal estimator for the objective function of peak signal-to-noise ratio (PSNR). This estimator is numerically determined by using variational Bayes (VB). We then solve the conjugate prior problem on VB and the exponential-order calculation cost problem of a compound Gaussian MRF prior with simple Taylor approximations. In experiments, the proposed method roughly overcomes existing methods.
\end{abstract}
\section{INTRODUCTION}
Super-resolution (SR) is a promising technology that is expected to be applied to microscope time series images, satellite photographs, and so on. SR aims at reconstructing a spatially high-resolution (HR) image from multiple low-resolution (LR) images. When multiple LR images obtained for the same object have transformation between images, they will contain mutually complementary information, and it becomes possible to estimate the HR image. Since the earliest work \cite{Tsai1984}, SR has been realized through various methods. 

In this work, we handle SR in a Bayesian framework \cite{Hardie1997,Tipping2003}. In a Bayesian framework, what prior we use is quite important. For example, various Markov random field (MRF) priors \cite{Geman1984,Molina2003a,Kanemura2007,Kanemura2009,Chellappa1991,Jeng1991,Katsuki2011}, the total variation (TV) prior \cite{Villena2010,Babacan2011}, the Huber prior \cite{Pickup2007}, and patch-based priors \cite{Ranzato2010} have been used in image processing. These can represent image properties well and have good performance in SR, image restoration, and other applications.

As the SR estimator, we think posterior mean (PM) is suitable as an HR image estimator because we usually evaluate the accuracy of SR methods by L2-norm (mean square error) -based peak signal-to-noise ratio (PSNR), and PM is the optimal estimator when employing PSNR as the objective function. To determine the exact PM of the HR image, all parameters other than the HR image should be marginalized out over the joint posterior distribution without using any point estimation. According to this meaning, the previous methods \cite{Tipping2003,Hardie1997,Kanemura2007,Kanemura2009,Babacan2011,Villena2010} are not optimal. Recently, the PM approach was proposed \cite{Katsuki2011}. It was the first method to employ the PM as the HR image estimator and does not use any point estimation of the registration parameters or the model parameters.

In this manuscript, we propose a new SR method that employs a compound Gaussian MRF prior that can simultaneously represent smoothness and edge of image \cite{Geman1984} and utilizes variational Bayes to calculate the optimal estimator, PM, with respect to the objective function of the PSNR. This approach seems quite favorable, but possibly it was not proposed earlier because of an important limitation of variational Bayes that a conjugate prior is needed, and because of the exponential-order calculation cost for a compound Gaussian MRF prior. We solve these problems through simple Taylor approximations. In Section 2, we show the formulation regarding the models and the estimator. In Section 3, we evaluate the proposed method. In Section 4, we discuss.

\section{FORMULATION}
\subsection{Notation}
First, we show the definitions of the gamma, Bernoulli and Gaussian distributions, the Kullback-Leibler (KL) divergence from distribution $p(\bx)$ to $q(\bx)$, and PSNR, used in this manuscript:
\begin{align}
	&\GammaDist(x; a, b)
	\equiv \frac{b^a}{\Gamma(a)} x^{a-1} \rme^{-b x} \quad (x > 0),\nonumber\\
	&\BernoulliDist(x; \mu)
	\equiv \mu^x (1-\mu)^{1-x} \quad (x \in \{ 0, 1 \}), \nonumber\\
	&\NormalDist(\bx; \bmu, \bSigma)
	\equiv |2\rmpi\bSigma|^{-\frac{1}{2}} \rme^{-\frac{1}{2} (\bx-\bmu)^\top \bSigma^{-1} (\bx-\bmu)} \quad (\bx \in \mathcal{R}^d), \nonumber\\
    &D_\mathrm{KL}(p(\bx) \| q(\bx))
	\equiv \left\langle \ln p(\bx)-\ln q(\bx) \right\rangle_{p(\bx)}, \nonumber\\
	&\mathrm{PSNR}(\hbx; \bx) \equiv 10 \log_{10} \frac{2^2}{\frac{1}{d} \| \hbx - \bx \|_2^2}.\nonumber
\end{align}
Here, $\Gamma$ is the gamma function, $|\bullet|$ denotes the determinant of a given matrix, $d$ is the dimension of $\bx$, and the angle brackets $\langle\bullet\rangle_\circ$ denote the expectation of $\bullet$ with respect to a distribution $\circ$. Also, $\mathrm{diag}$ denotes a diagonal matrix. All the vectors in this manuscript are column vectors. Here, these variables have absolutely nothing to do with the variables that appear later.

\subsection{Observation model}
We estimate an HR grayscale image $\bx \in \mathcal{R}^{N_\bx}$ from observed multiple LR grayscale images $\bY \equiv \{\by_l \}_{l=1}^L, \by_l \in \mathcal{R}^{N_\by}$ using an SR technique. The images $\by_l$ and $\bx$ are regarded as lexicographically stacked vectors. The number of pixels for each LR image is assumed to be less than that of the HR image; i.e., $N_\by < N_\bx$. Although we define the range of a pixel luminance value as infinite, we use $-1$ for black, $+1$ for white, and values between $-1$ and $+1$ for shades of gray. The HR image $\bx$ is geometrically warped, blurred, downsampled, and corrupted by noise $\bepsilon_l$ to form the observed LR image $\by_l$. It is modeled as
\begin{align}
	\label{EqImageObservationProcess}
	&\by_l\! \equiv \!\bW(\!\bphi_l\!)\bx \!+\! \bepsilon_l,
	~~~~p(\!\bY | \bx\!,\! \beta\!,\! \bPhi\!)
	\!\equiv\! \prod_{l=1}^L \!\NormalDist\!(\!\by_l ; \bW(\!\bphi_l\!) \bx, \beta^{-1}\!\bI\!),\\
	&\bPhi \equiv \{ \bphi_l \}_{l=1}^L, ~~~~
	\bphi_l
	\equiv [ \phi_{l,k} ]_{k=1}^4
	\equiv [ \theta_l, [\vec{o}_l]_x, [\vec{o}_l]_y, \gamma_l ]^\top.
\end{align}
The $\bepsilon_l \in \mathcal{R}^{N_\by}$ is additive white Gaussian noise (AWGN) with precision (inverse variance) $\beta$ ($>0$). $\bW(\bphi_l)$ is the transformation matrix that simultaneously applies warping, blurring, and downsampling. The details of this matrix are given in \cite{Katsuki2011}. $\bphi_l$ is a four-dimensional vector consisting of the registration parameters: rotational motion parameter $\theta_l$, translational motion parameter $\vec{o}_l$, and blurring parameter $\gamma_l$.

\subsection{Prior Distributions}
We use a compound Gaussian MRF prior for the HR image and the latent variables $\bmeta$ representing the edges, called a line process, that is known to be favorable for natural images. It is a compounded distribution of the Gaussian MRF model and the line process proposed by \cite{Geman1984}, which is widely used \cite{Molina2003a,Chellappa1991,Jeng1991} and can simultaneously represent smoothness and discontinuity of the image. It is defined as
\begin{align}
	\label{EqPriorDistributionAnotherXEta}
	&p(\bx, \bmeta|\lambda,\rho,\kappa)\nonumber\\
	&\equiv\! \frac{\!\exp\!\bigg[\!-\!\lambda \!\sum_{i \sim j} (\!1 \!-\! \eta_{i,j}\!) \!-\! \frac{\rho}{2} \!\sum_{i\!\sim\! j}\!\eta_{i,j}(\!x_i\!-\!x_j\!)^2\! - \!\frac{\kappa}{2} \!\|\bx\|_2^2\!\bigg]}{\!\sum_\bmeta \!\int\! \exp\!\bigg[\!-\!\lambda \!\sum_{i \!\sim\! j}\! (\!1 \!-\! \eta_{i,j}\!)\! - \!\frac{\rho}{2} \!\sum_{i\!\sim\! j}\!\eta_{i,j}\!(\!x_i\!-\!x_j\!)^2 \!-\! \frac{\kappa}{2} \!\|\bx\|_2^2\!\bigg] \!\mathrm{d}\bx}\nonumber\\
    &=\exp\bigg[-\lambda \sum_{i \sim j} (1 \!-\! \eta_{i,j}) - \frac{1}{2}\bx^\top\bA(\bmeta, \rho, \kappa)\bx\nonumber\\ 
    &~~~~\!-\!\ln\!\sum_{\bmeta}\!\exp\bigg\{\!-\!\lambda\sum_{i\sim j}(1\!-\!\eta_{i,j})\!-\!\frac{1}{2}\ln\!\bigg|\frac{1}{2\pi}\bA(\bmeta, \rho, \kappa)\bigg|\bigg\}\bigg],\\
	\label{EqAMatrix}
	&~~~~~~~~\bA(\bmeta, \rho, \kappa)_{i,j}
	\equiv
	\begin{cases}
		\rho \sum_{k \sim i} \eta_{i,k} + \kappa, & i=j, \\
		-\rho \eta_{i,j},                         & i \sim j, \\
		0,                                        & \mathrm{otherwise}.
	\end{cases}
\end{align}
The summation $\sum_{i \sim j}$ is taken over all pairs of adjacent pixels. The notation $i \sim j$ means that the $i$-th and the $j$-th pixels are adjacent in upward, downward, leftward, or rightward directions. The line process $\bmeta$ consists of binary latent variables $\eta_{i,j} \in \{0,1 \}$ for all adjacent pixel pairs $i$ and $j$. Its size equals $N_\bmeta \equiv 2 N_\bx - [\textrm{number of HR image's horizontal pixels}] - [\textrm{number of HR image's vertical pixels}]$. The hyperparameter $\lambda$ ($>0$) is an edge-penalty parameter which prevents $\eta_{i,j}$ from excessively taking edges. Also, $\rho$ ($>0$) is a smoothness parameter which prevents differences in adjacent pixel luminance from becoming large, and $\kappa$ ($>0$) is a contrast parameter which prevents $\bx$ from taking an improperly large absolute value.

Here, the ``causal'' Gaussian MRF prior used in \cite{Kanemura2007,Kanemura2009,Katsuki2011} is defined as the joint distribution of $\bx$ and $\bmeta$ in the form of $p(\bmeta)p(\bx|\bmeta)$, and it differs from the compound one in that it is not simultaneously normalized about both $\bx$ and $\bmeta$ like Eq. (\ref{EqPriorDistributionAnotherXEta}). A ``causal'' one is an approximation of the compound one, and it is easier to use than the compound one because simultaneous normalization of a compound one has an exponential-order calculation cost with respect to the dimensionality of the line process; the calculation cost of a ``causal'' one is polynomial. Additionally, though both $\bx$ and $\bmeta$ of a compound one has a Markov property, in the ``causal'' one only $\bx$ has a Markov property. However, in Eq. (\ref{EqPriorDistributionAnotherXEta}), ignoring $\ln|\bA|$ as in \cite{Kanemura2007,Kanemura2009} makes them take the same form and breaks either property. Therefore, in Section 3, we propose a new approximation that does not ignore $\ln|\bA|$.

The hyperparameter priors and the registration parameter priors are defined as
\begin{align}
\label{EqPriorDistributionLambda}
	p(\lambda, \rho, \kappa, \beta) &\equiv
	\GammaDist(\lambda; a_\lambda^\stepOO, b_\lambda^\stepOO)
	\GammaDist(\rho; a_\rho^\stepOO, b_\rho^\stepOO) \nonumber\\
	&\times \GammaDist(\kappa; a_\kappa^\stepOO, b_\kappa^\stepOO)
	\GammaDist(\beta; a_\beta^\stepOO, b_\beta^\stepOO),\\
	\label{EqPriorDistributionPhi}
	p(\bPhi) &\equiv \prod_{l=1}^L \NormalDist(\bphi_l; \bmu_{\bphi_l}^\stepOO, \bSigma_{\bphi_l}^\stepOO),
\end{align}
where $a_\lambda^\stepOO \!\equiv\! 3\times10^{-2}$, $b_\lambda^\stepOO, a_\rho ^\stepOO,b_\rho ^\stepOO ,a_\kappa ^\stepOO,b_\kappa ^\stepOO ,a_\beta ^\stepOO,b_\beta ^\stepOO \!\equiv\! 10^{-2}$, $\bmu_{\bphi_l}^\stepOO \!\equiv\! [0, 0, 0, 12/\alpha^2]$ and $ \bSigma_{\bphi_l}^\stepOO\! \equiv\! \mathrm{diag}[10^{-3}, 10^{0}, 10^{0}, 10^{-3}]$. For a gamma distribution, the number of effective prior observations in the Bayesian framework is equal to two times parameter $a$. The above settings are considered sufficiently non-informative. Here, we assume $\lambda > 0$ by setting its prior according to a gamma distribution similar to \cite{Katsuki2011}, resulting in an appropriate inference. We define the mean value for the gamma distribution as $\mu_\lambda^\steptt \!\equiv\! \frac{a_\lambda^\steptt}{b_\lambda^\steptt}, \mu_\rho^\steptt \!\equiv\! \frac{a_\rho^\steptt}{b_\rho^\steptt}$, $\mu_\kappa^\steptt \!\equiv\! \frac{a_\kappa^\steptt}{b_\kappa^\steptt}$, and $\mu_\beta^\steptt \!\equiv\! \frac{a_\beta^\steptt}{b_\beta^\steptt}$. $t$ denotes an iterative step later introduced through variational Bayes. The settings of the registration parameter priors are considered suitable for this SR task \cite{Katsuki2011}. Note that the mean value $\mu_{\gamma_l}^\stepOO$ of the $\gamma_l$ prior is derived as the value equivalent to the anti-aliasing of the scale factor \cite{Katsuki2011}.

\subsection{Objective function and optimal estimator}
First, we confirm that the joint distribution of all random variables can now be explicitly given as
\begin{align}
	p(\bY, \bz)
	&= p(\bY | \bx, \beta, \bPhi) p(\bx, \bmeta | \lambda, \rho, \kappa) p(\lambda, \rho, \kappa, \beta) p(\bPhi), \\
	\bz
	&\equiv [ \bx, \bmeta, [\lambda, \rho, \kappa, \beta], \bPhi ].
\end{align}
We define the objective function using $\mathrm{PSNR}(\hbx(\bY); \bx) $ and optimal estimator as
\begin{align}
	\label{ObjectiveFunction}
	&\argmax_{\hbx(\bY)} \left[10 \log_{10} \frac{2^2}{\left\langle\frac{1}{N_{\bm{x}}} \| \hat{\bm{x}}(\bm{Y}) - \bm{x} \|_2^2\right\rangle_{p(\bm{Y},\bm{x})}}\right]
	= \left\langle \bx \right\rangle_{p(\bx | \bY)}.
\end{align}
Since only LR images, $\bY$, are available for the estimator, we sometimes explicitly express it as a function form, $\hbx(\bY)$. We choose this objective function because we prefer good estimator performance on average over various HR images and the corresponding LR images. From Eq. (\ref{ObjectiveFunction}), the PM $\langle\bx \rangle_{p(\bx | \bY)}$ is the best estimator of the HR image. Note that the posterior distribution of an HR image $p(\bx | \bY)$ needs marginalization of all parameters other than $\bx$ over $p(\bz | \bY)$.

\subsection{Approximation methods}
As stated above, we could derive the optimal estimator. However, we cannot obtain the marginalized posterior distribution $p(\bx | \bY)$ analytically. Therefore, we solve this through approximation by using variational Bayes \cite{Attias1999}. We impose the factorization assumption on the trial distribution $q(\bz)\!\equiv\! q(\bx) q(\bmeta) q(\lambda, \!\rho, \!\kappa, \!\beta) q(\bPhi)$. The optimal trial distribution is identified by minimizing the KL divergence between the trial and the true distributions as the best approximation of the true distribution: $\hq(\bz)\!\equiv\! \argmin_{q(\bz)} D_\mathrm{KL}(q(\bz) \| p(\bz | \bY))$. In the common style of variational Bayes \cite{Bishop2003,Babacan2011}, update equations are
\begin{align}
	\label{EqRecurrenceInitiation}
	q^\stepOO(z_i)
	\equiv p(z_i), ~~~~~~q^\steptl(z_i)\!\propto \!\exp \!\langle \ln \!p(\bz | \bY) \rangle\!_{ \prod_{j \neq i}\! q^\steptt(z_j)}.
\end{align}
Frequently, the application of variational Bayes is difficult in practice because it requires a conjugate prior. We make the priors conjugate by using simple Taylor approximations similar to \cite{Katsuki2011}. We also solve the exponential-order calculation cost problem of a compound Gaussian MRF prior by using the same approach here. These approximations enable the analytical calculations in Eq. (\ref{EqRecurrenceInitiation}). Specifically, we apply the first-order Taylor approximations for three non-linear terms. First, in Eq. (\ref{EqImageObservationProcess}), $\bW(\bphi_l)$ is approximated around $\bphi_l = \bmu_{\bphi_l}^\steptt$ the same as in \cite{Katsuki2011,Babacan2011,Villena2010}. Second, the logarithm of the normalization term of Eq. (\ref{EqPriorDistributionAnotherXEta}) is approximated around $\ln \lambda = \ln \mu_\lambda^\steptt$. Finally, in Eq. (\ref{EqPriorDistributionAnotherXEta}), $\ln\left|\bA(\bmeta, \rho, \kappa)\right|$ is approximated around $[\bmeta, \ln \rho, \ln \kappa] = [\bmu_\bmeta^\steptl, \ln \mu_\rho^\steptt, \ln \mu_\kappa^\steptt]$ similar to \cite{Katsuki2011}. Note that in \cite{Katsuki2011}, the third approximation is the key idea to solve the conjugate prior problem. In this work, we found it also enables us to solve the exponential-order calculation cost problem of a compound Gaussian MRF prior. This makes it possible to calculate the normalization term.

\subsection{Algorithm}
From Eq. (\ref{EqRecurrenceInitiation}) and the Taylor approximations, the trial distributions are obtained as the following distributions:
\begin{align}
	\label{EqTrialDistributionEtaX}
	&q^\steptt(\bmeta)
	\!=\! \prod_{i \sim j} \!\BernoulliDist(\!\eta_{i,j}; \mu_{\eta_{i,j}}^\steptt\!), 
	~~q^\steptt(\bx)
	\!=\! \NormalDist\!(\!\bx; \bmu_\bx^\steptt, \!\bSigma_\bx^\steptt\!), \\
	\label{EqTrialDistributionLambda}
	&q^\steptt(\lambda, \rho, \kappa, \beta)
	= \GammaDist(\lambda; a_\lambda^\steptt, b_\lambda^\steptt)
	   \GammaDist(\rho;    a_\rho   ^\steptt, b_\rho   ^\steptt) \nonumber\\
	&~~~~~~~~~~~~~~~~~~~~~~~\times
	   \GammaDist(\kappa;  a_\kappa ^\steptt, b_\kappa ^\steptt)
	   \GammaDist(\beta;   a_\beta  ^\steptt, b_\beta  ^\steptt), \\
	\label{EqTrialDistributionPhi}
	&q^\steptt(\bPhi)
	= \prod_{l=1}^L \NormalDist(\bphi_l; \bmu_{\bphi_l}^\steptt, \bSigma_{\bphi_l}^\steptt).
\end{align}
The specific form of the trial distributions is omitted because of space limitations. For Eq. (\ref{EqRecurrenceInitiation}), we do the following. First, we compute $q^\steptl(\bmeta)$ using $q^\steptt(\bx, \lambda, \rho, \kappa, \beta, \bPhi)$. Second, we compute $q^\steptl(\bx)$ using $q^\steptl(\bmeta) q^\steptt(\lambda, \rho, \kappa, \beta, \bPhi)$. Last, we compute $q^\steptl(\lambda, \rho, \kappa, \beta)$ using $q^\steptl(\bx, \bmeta) q^\steptt(\bPhi)$ and $q^\steptl(\bPhi)$ using $q^\steptl(\bx, \bmeta) q^\steptt(\lambda, \rho, \kappa, \beta)$. For the initial parameters of the trial distributions of $\bmeta$ and $\bx$, we use non-informative values, $\bmu_\bmeta^\stepOO \equiv \bm{0},~ \bmu_\bx^\stepOO \equiv \bm{0},~ \bSigma_\bx^\stepOO \equiv \bm{0}$. As the initial parameters for $\lambda$, $\rho$, $\beta$, $\kappa$, and $\bPhi$ we use the same values as their prior's values in Eq. (\ref{EqPriorDistributionLambda}), (\ref{EqPriorDistributionPhi}). We obtain the well approximated PM of $\bx$ as $\bmu_\bx^\steptl$, for which the following convergence conditions hold for $\bmu_\bx^\steptl$ and each $\mu_{\phi_{l,k}}^\steptl$,
\begin{align}
	\frac{1}{N_\bx} \| \bmu_\bx^\steptl - \bmu_\bx^\steptt \|_2^2 &< 10^{-5},\nonumber\\
	\frac{1}{L} \sum_{l=1}^L \frac{(\mu_{\phi_{l,k}}^\steptl-\mu_{\phi_{l,k}}^\steptt )^2}{[\bsigma^2_{\bphi}]_{k}} &< 10^{-5} \quad (k=1,2,3,4),
\end{align}
where we defined $\bsigma^2_{\bphi} \equiv [10^{-3}, 10^{0}, 10^{0}, 10^{-3}]$ as the scaling constant.

\section{EXPERIMENTAL RESULTS}
\begin{figure}[tb]
\centering
	{\includegraphics[width=15mm]{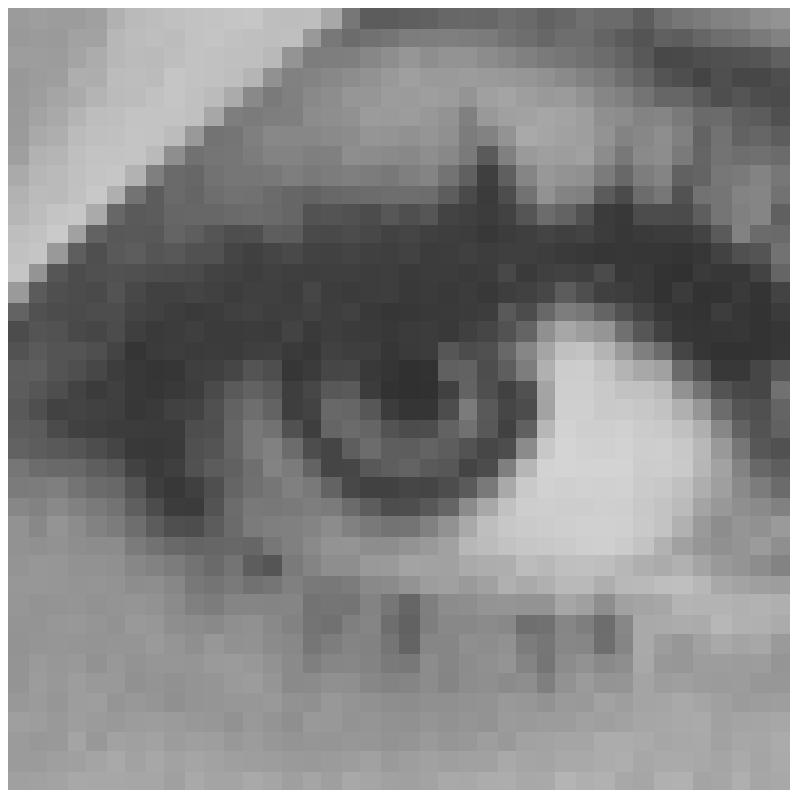}
	\label{OriginalLena}}
	{\includegraphics[width=15mm]{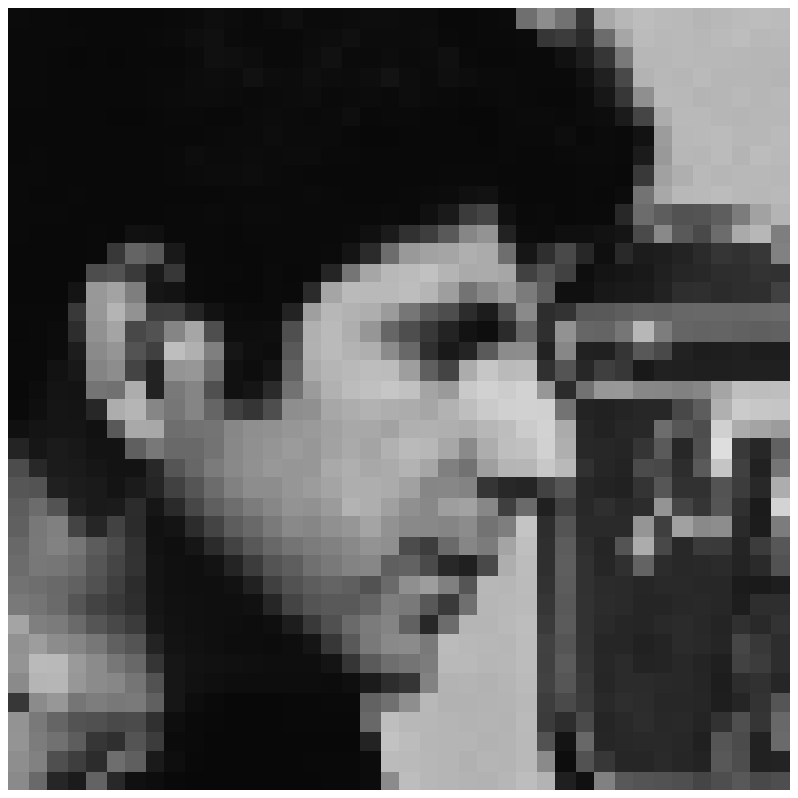}
	\label{OriginalCameraman}}
	{\includegraphics[width=15mm]{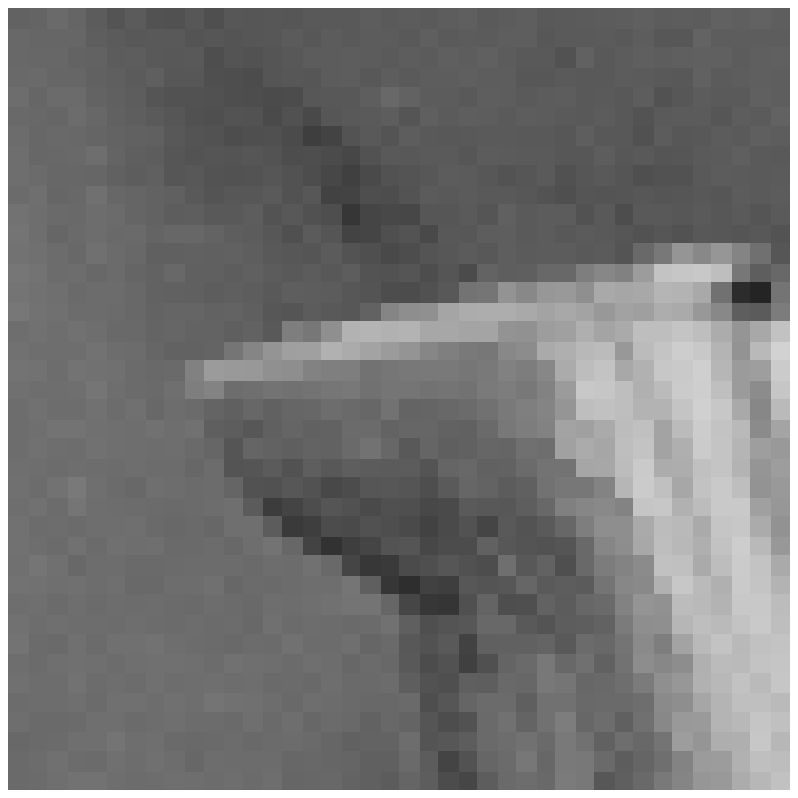}
	\label{OriginalPepper}}
	{\includegraphics[width=15mm]{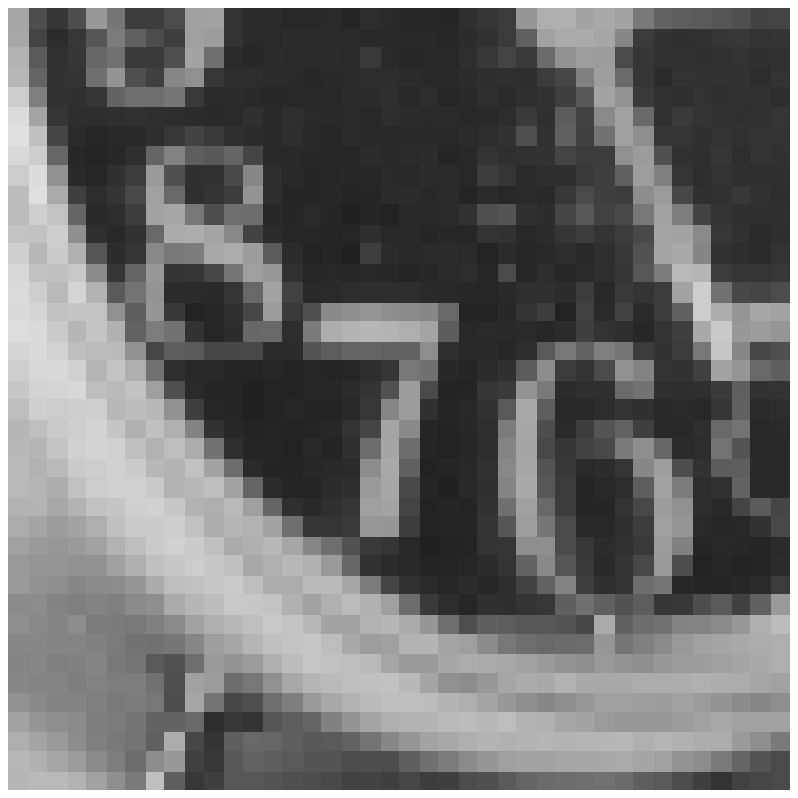}
	\label{OriginalClock}}
	{\includegraphics[width=15mm]{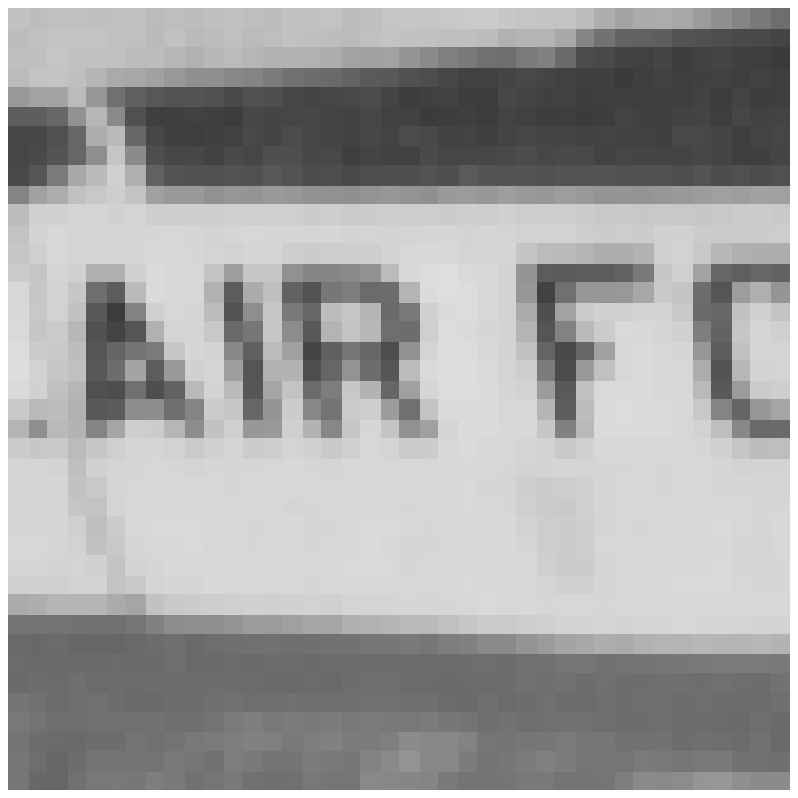}\vspace{-8pt}
	\label{OriginalText}}
\caption{Five original images used in the experiments. From left to right: Lena, Cameraman, Pepper, Clock, Text.}\vspace{-6pt}
	\label{FigOriginal}
\end{figure}

\begin{figure}[t]
\centering
	{\includegraphics[width=15mm]{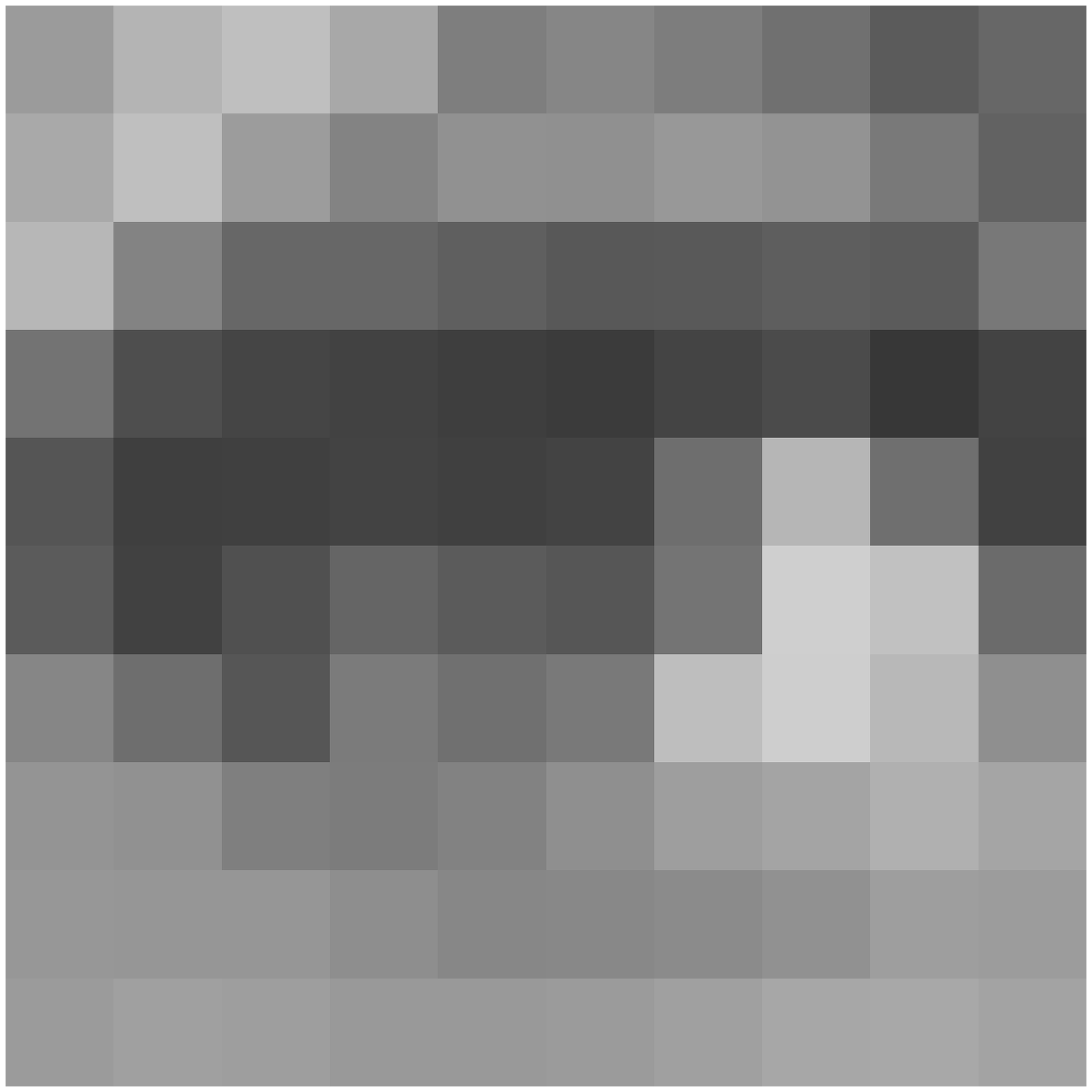}
	\label{ObservedLena}}
	{\includegraphics[width=15mm]{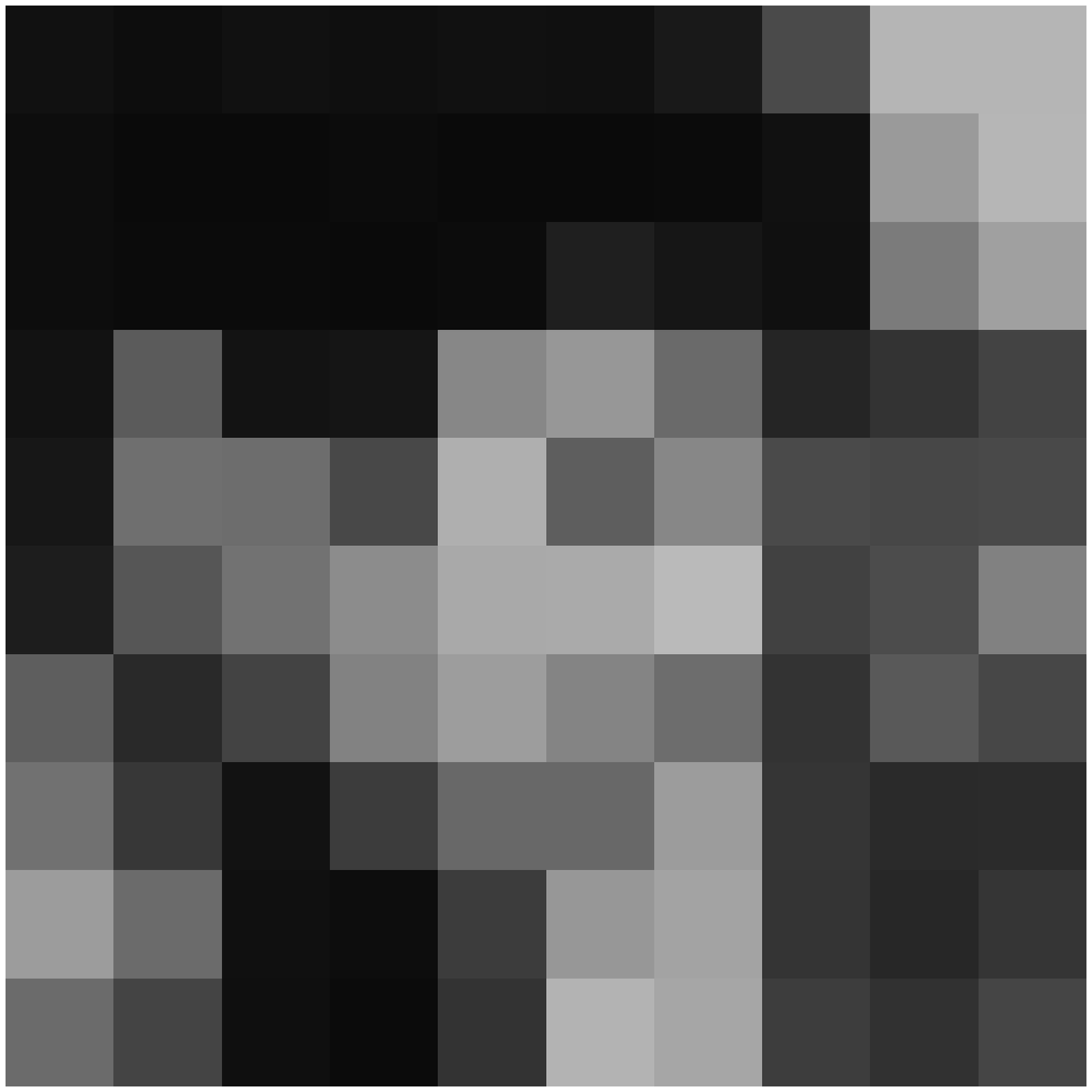}
	\label{ObservedCameraman}}
	{\includegraphics[width=15mm]{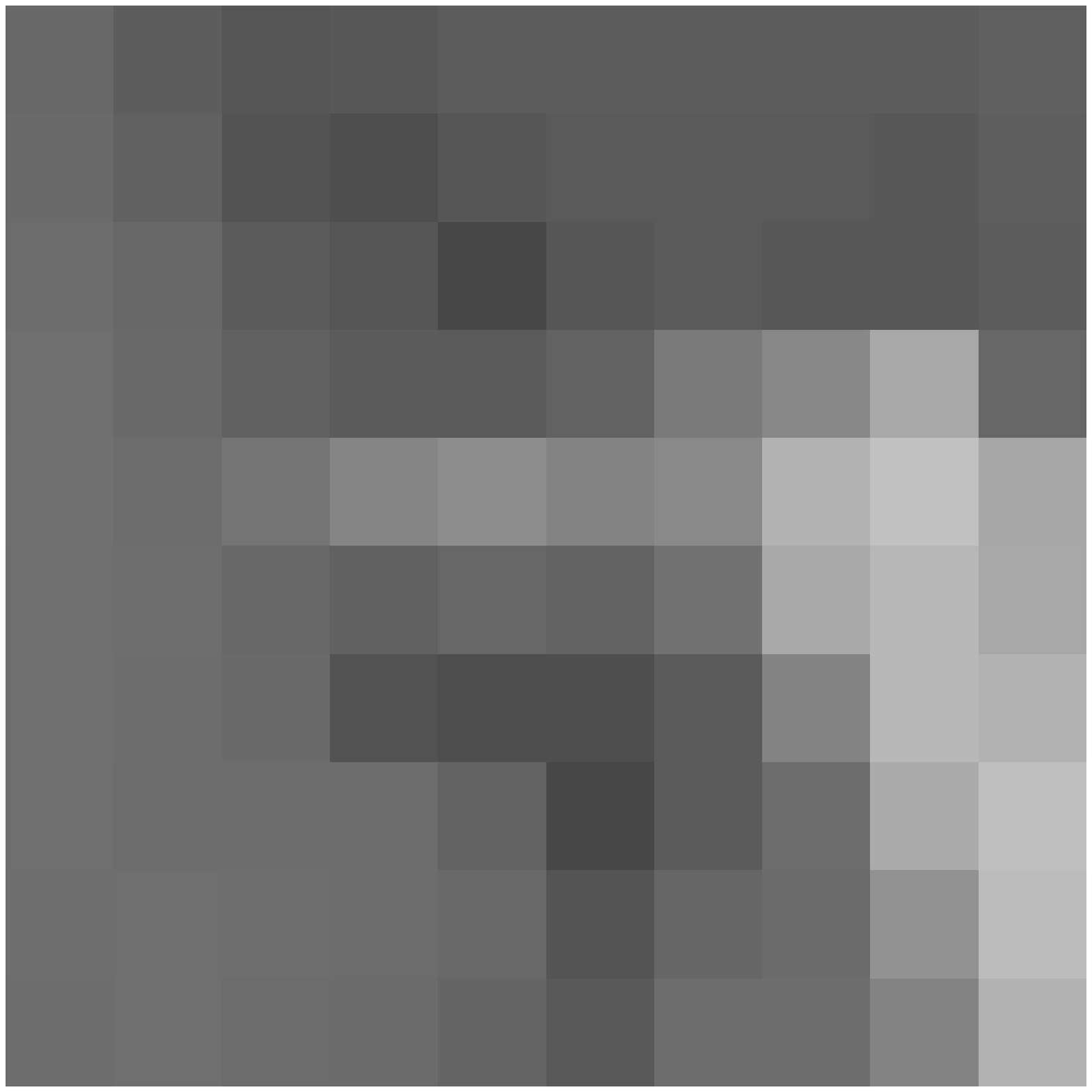}
	\label{ObservedPepper}}
	{\includegraphics[width=15mm]{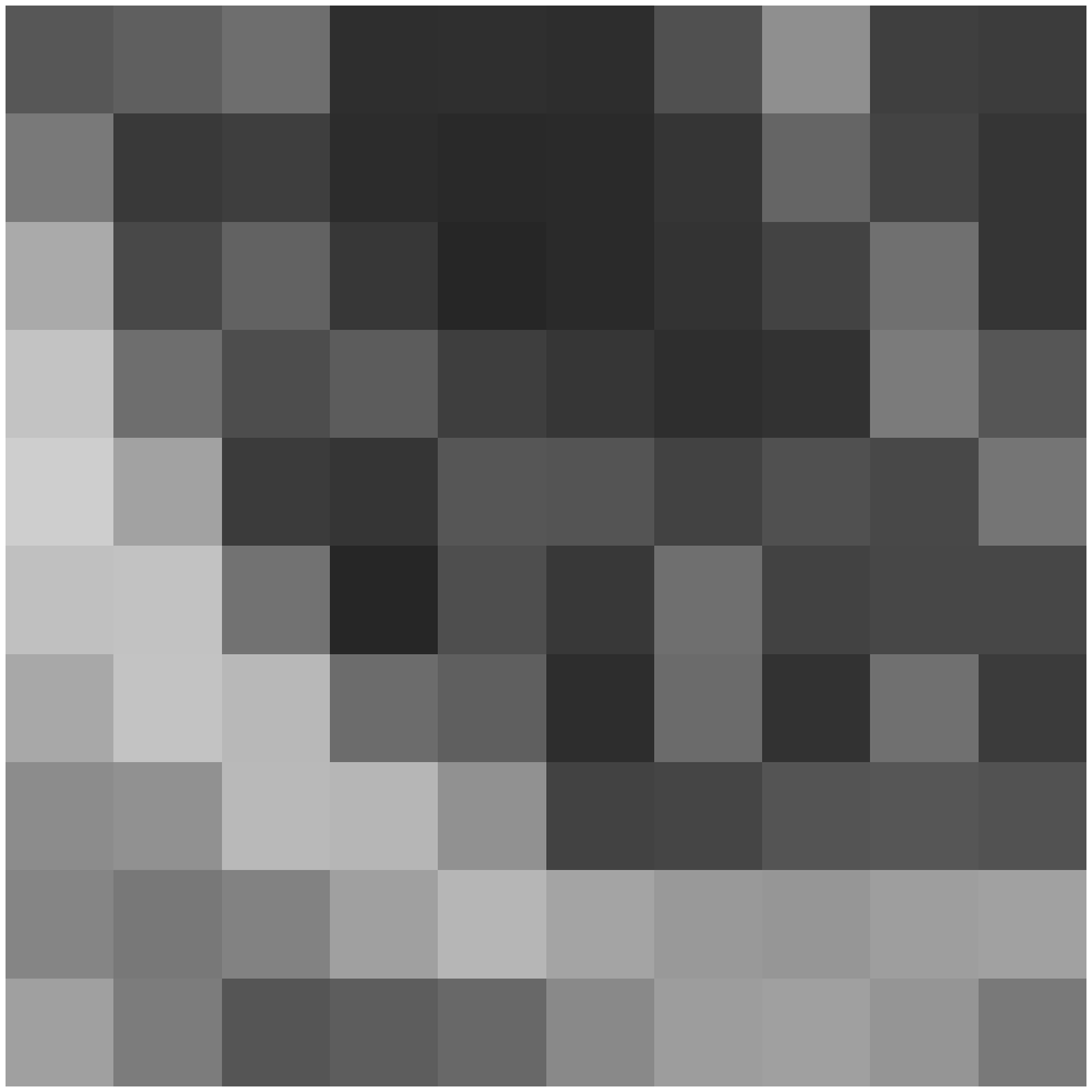}
	\label{ObservedClock}}
	{\includegraphics[width=15mm]{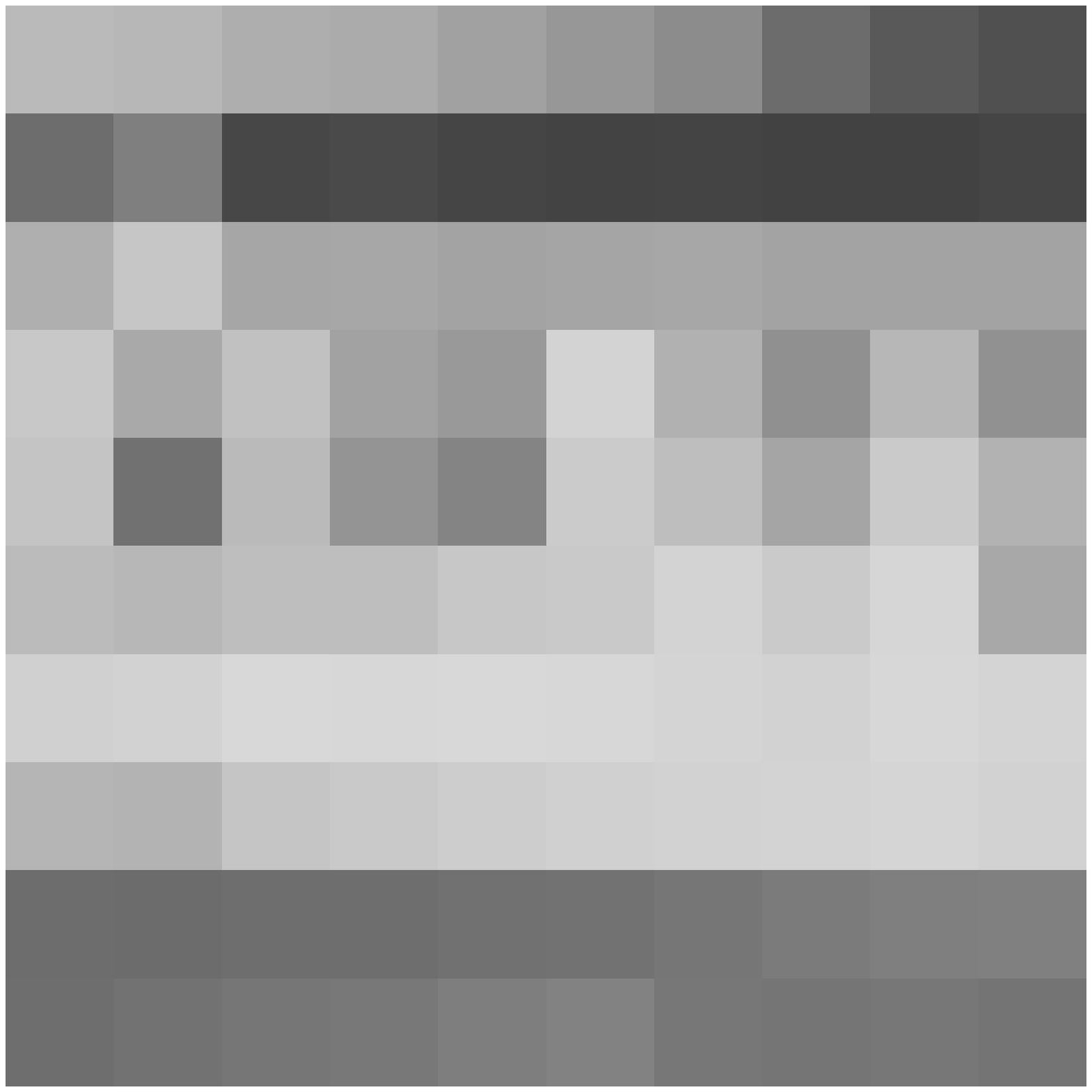}\vspace{-8pt}
	\label{ObservedText}}
\caption{Observed images when warped, blurred, downsampled by an enhancement factor of 4, and noised with SNR$=30$ dB AWGN}\vspace{-6pt}
	\label{FigObserved}
\end{figure}

\begin{figure}[t]
\centering
	{\includegraphics[width=15mm]{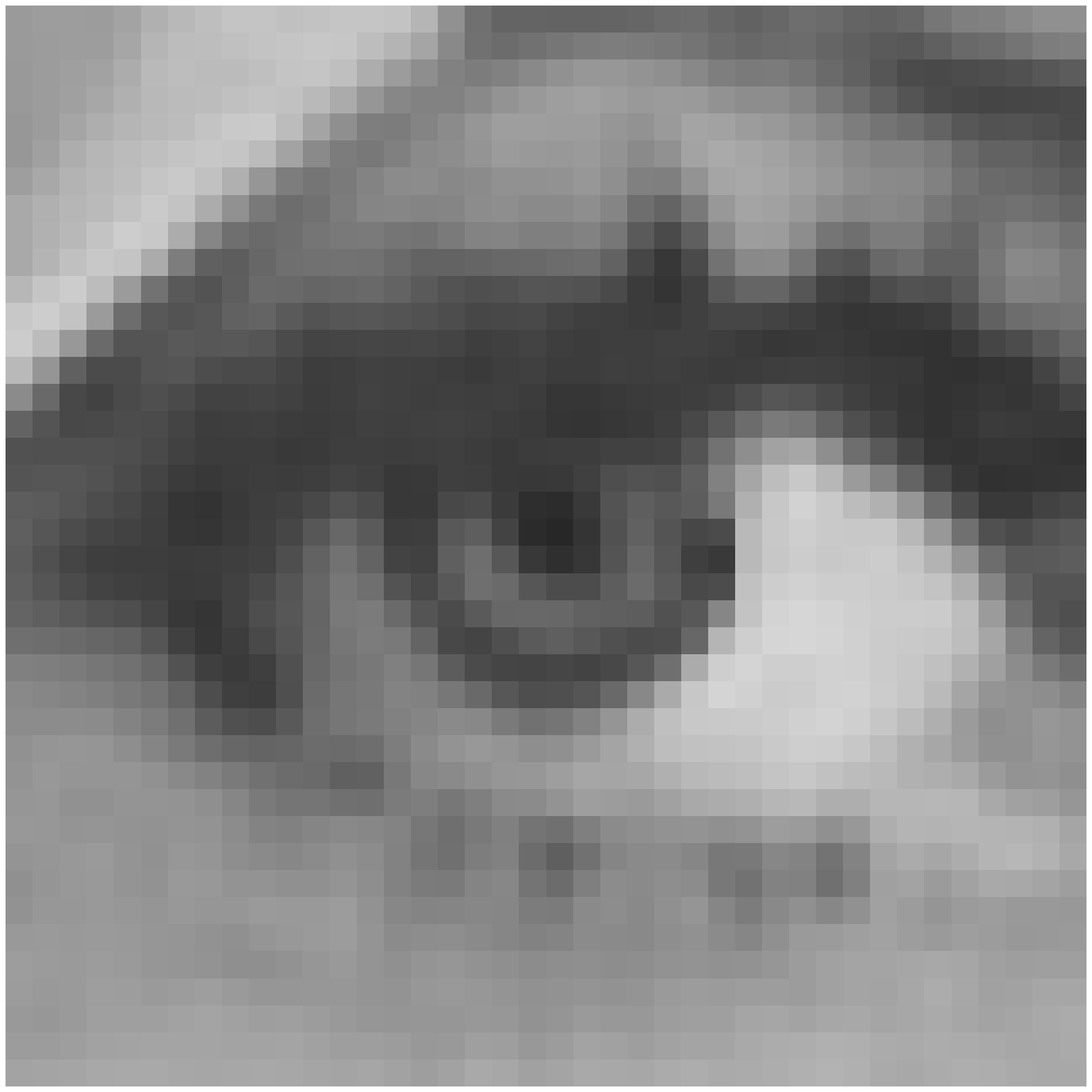}
	\label{InferredLena}}
	{\includegraphics[width=15mm]{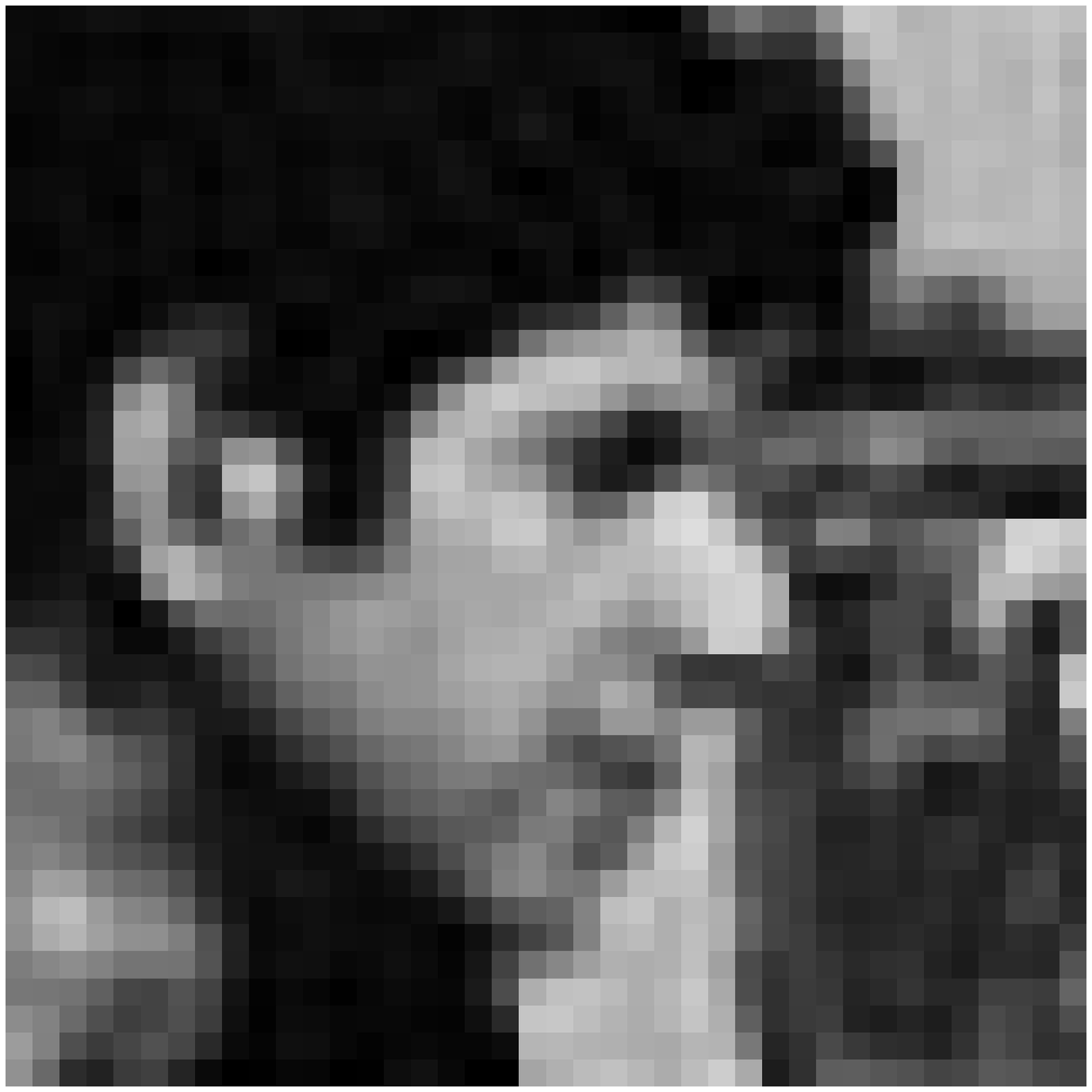}
	\label{InferredCameraman}}
	{\includegraphics[width=15mm]{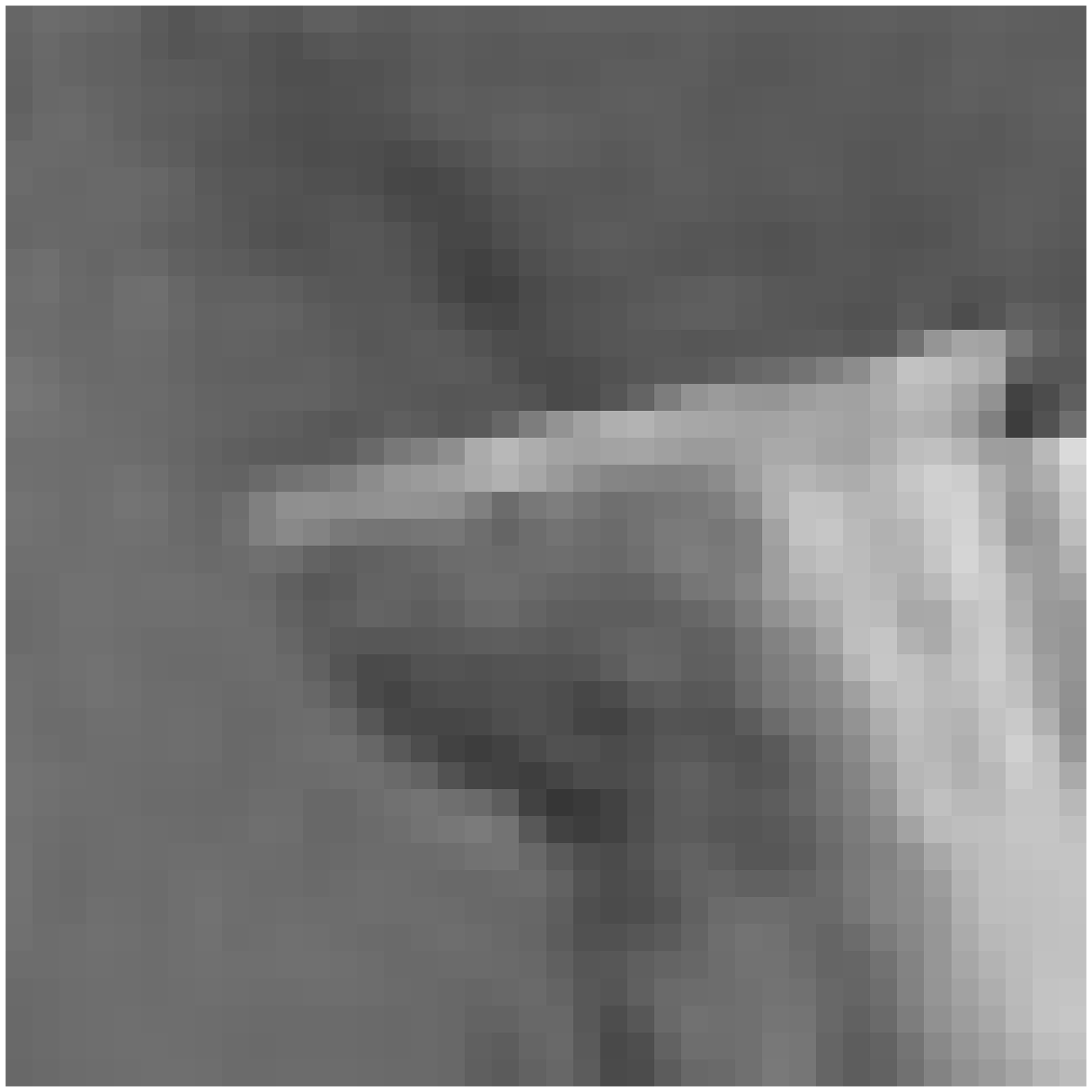}
	\label{InferredPepper}}
	{\includegraphics[width=15mm]{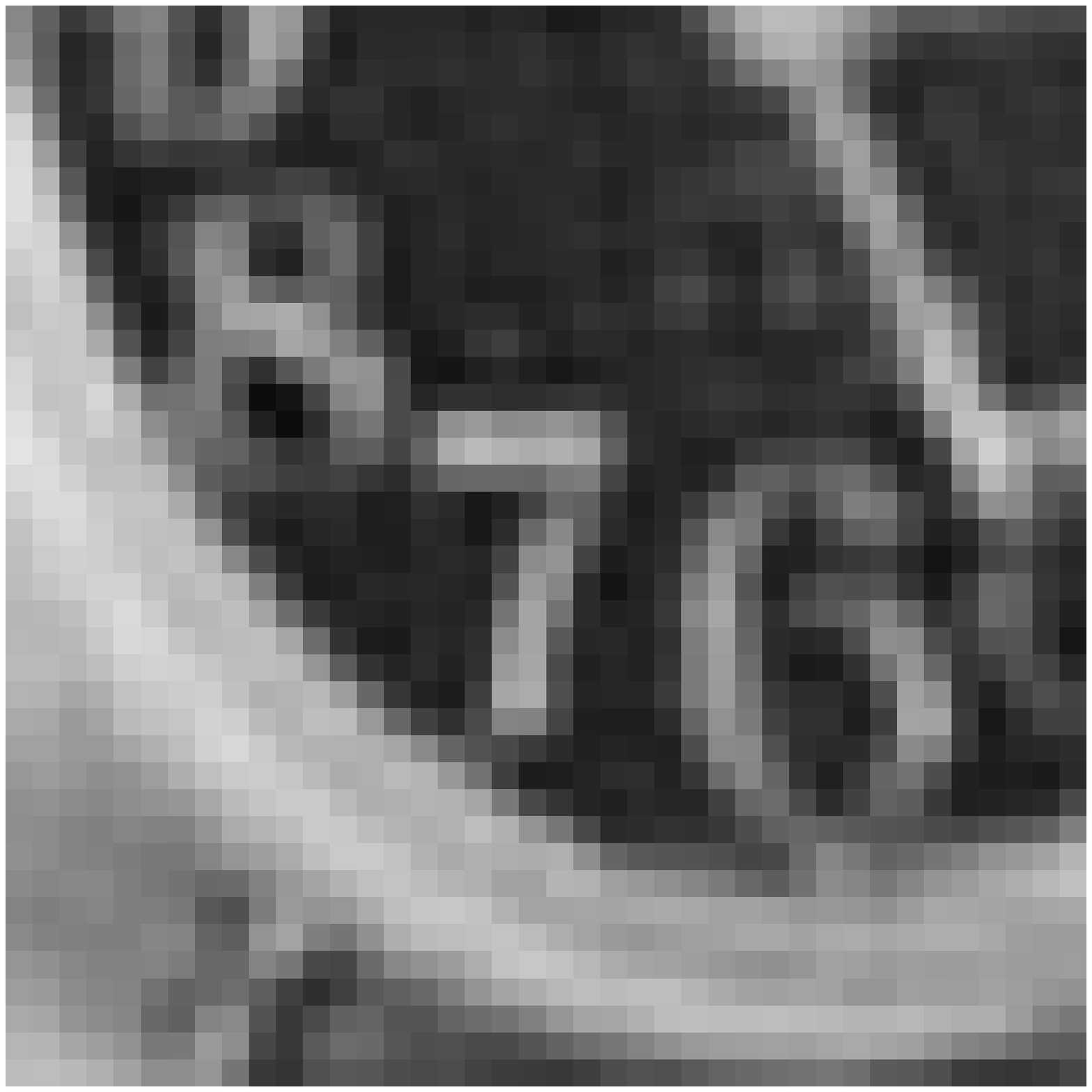}
	\label{InferredClock}}
	{\includegraphics[width=15mm]{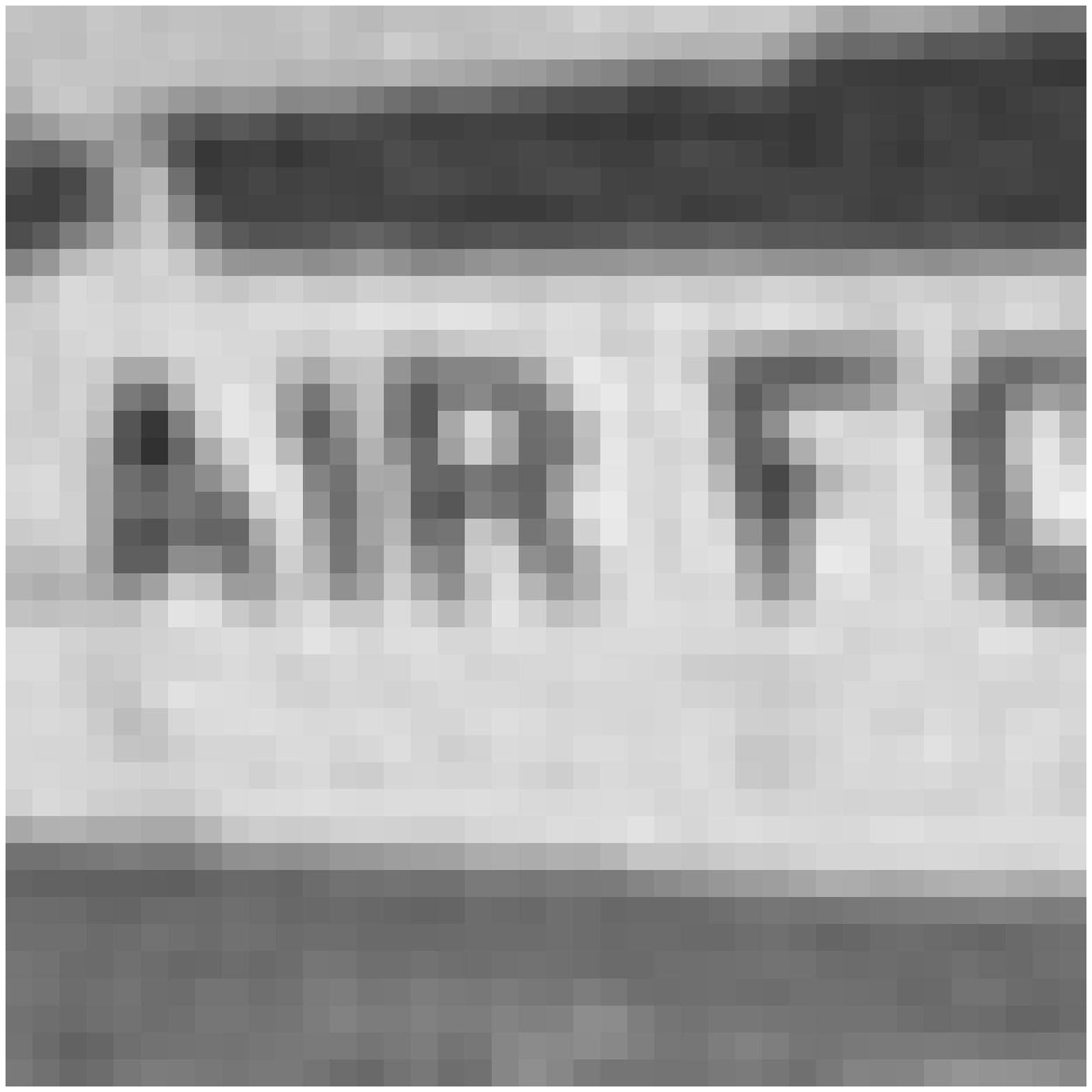}\vspace{-8pt}
	\label{InferredText}}
\caption{Images estimated from the Fig. \ref{FigObserved} observed images}\vspace{-6pt}
	\label{FigInferred}
\end{figure}

\begin{table*}[tb]
	\centering
	\caption{PSNR of the proposed method and ISNRs against four previous methods for different images and SNR levels}
	\label{TableResultsPSNR}
{

	\begin{tabular}{ccccccc}
\hline\hline
	Image & SNR [dB]& PSNR (proposed) & ISNR (a) & ISNR (b) & ISNR (c) & ISNR (d)\\
	\hline
	Lena & $20$ & $29.31 \pm 0.30$ & $ 5.45 \pm 0.33$ & $0.67 \pm 0.34$ & $ 0.02 \pm 0.11$ & $ 0.05 \pm 0.01$\\
	              & $30$ & $32.15 \pm 0.36$ & $ 8.20 \pm 0.37$ & $1.74 \pm 0.34$ & $ 0.52 \pm 0.18$ & $ 0.10 \pm 0.20$\\
	              & $40$ & $34.19 \pm 0.60$ & $ 10.24 \pm 0.60$ & $3.21 \pm 0.53$ & $ 0.95 \pm 0.60$ & $ 1.49 \pm 0.77$\\ \hline
	Cameraman & $20$ & $21.76 \pm 0.20$ & $ 4.13 \pm 0.21$ & $0.95 \pm 0.32$ & $ -0.04 \pm 0.08$ & $ -0.01 \pm 0.01$\\
	              & $30$ & $23.59 \pm 0.28$ & $ 5.92 \pm 0.28$ & $1.56 \pm 0.32$ & $ -0.01 \pm 0.11$ & $ -0.06 \pm 0.02$\\
	              & $40$ & $25.04 \pm 0.41$ & $ 7.37 \pm 0.42$ & $2.70 \pm 0.30$ & $ 0.32 \pm 0.27$ & $ -0.01 \pm 0.14$\\ \hline
	Pepper & $20$ & $29.73 \pm 0.24$ & $ 3.68 \pm 0.26$ & $0.09 \pm 0.41$ & $ 0.23 \pm 0.10$ & $ 0.11 \pm 0.87$\\
	              & $30$ & $31.65 \pm 0.33$ & $ 5.51 \pm 0.33$ & $0.76 \pm 0.48$ & $ 0.11 \pm 0.22$ & $ 0.35 \pm 0.33$\\
	              & $40$ & $32.23 \pm 0.51$ & $ 6.09 \pm 0.51$ & $1.11 \pm 0.45$ & $ -0.17 \pm 0.48$ & $ 0.93 \pm 0.56$\\ \hline
	Clock & $20$ & $23.29 \pm 0.28$ & $ 5.38 \pm 0.29$ & $1.40 \pm 0.23$ & $ 0.10 \pm 0.09$ & $ 0.01 \pm 0.01$\\
	              & $30$ & $25.42 \pm 0.29$ & $ 7.46 \pm 0.29$ & $2.59 \pm 0.30$ & $ 0.29 \pm 0.13$ & $ -0.03 \pm 0.01$\\
	              & $40$ & $27.08 \pm 0.38$ & $ 9.13 \pm 0.38$ & $4.00 \pm 0.31$ & $ 0.74 \pm 0.32$ & $ -0.07 \pm 0.12$\\ \hline
	Text & $20$ & $24.68 \pm 0.32$ & $ 5.83 \pm 0.33$ & $1.65 \pm 0.29$ & $ -0.06 \pm 0.09$ & $ 0.02 \pm 0.02$\\
	              & $30$ & $27.27 \pm 0.43$ & $ 8.38 \pm 0.44$ & $3.09 \pm 0.41$ & $ 0.19 \pm 0.18$ & $ -0.03 \pm 0.04$\\
	              & $40$ & $29.28 \pm 0.62$ & $ 10.39 \pm 0.62$ & $4.85 \pm 0.51$ & $ 0.78 \pm 0.51$ & $ 1.98 \pm 0.69$\\
	\hline\hline
\end{tabular}
}
\end{table*}
We evaluated the proposed method using five gray-scale images with a size of $40 \times 40$ pixels as shown in Fig. \ref{FigOriginal}. From each image, $L=10$ images with a size of $10 \times 10$ pixels were created by using Eq. (\ref{EqImageObservationProcess}). The resolution enhancement factor was 4. The transformation parameter $\bPhi$ was generated from the prior distribution in Eq. (\ref{EqPriorDistributionPhi}), where it is similar to that in previous work \cite{Tipping2003,Kanemura2007,Kanemura2009,Babacan2011,Villena2010,Katsuki2011,Pickup2007}. The noise level parameter $\beta$ was set for a signal-to-noise ratio (SNR) of $20$, $25$, and $30$ dB for each image. Samples of the created images are shown in Fig. \ref{FigObserved}. Figure \ref{FigInferred} shows the estimated images under SNR$=30$dB. We can see that the resolution of each image appeared to be better than that of the corresponding observed image.

Table \ref{TableResultsPSNR} shows the quantitative results compared to methods (a), (b), (c), and (d), where (a) is bilinear interpolation, (b) is the variational EM approach with a causal Gaussian MRF prior using the maximum a posteriori (MAP) estimator \cite{Kanemura2007}, (c) is the variational Bayes approach with a TV prior using the MAP estimator \cite{Babacan2011}, and (d) is the PM approach with a causal Gaussian MRF prior using the PM estimator \cite{Katsuki2011}. Note that we added a slight modification to methods (b) and (c), the same as in \cite{Katsuki2011}, because they employ slightly different models. Over $100$ experiments on each image and for each SNR, we evaluated the results with regard to the expectation and the standard deviation of the PSNR of the proposed method and the improvement in signal-to-noise ratio (ISNR) defined as, 
$
\mathrm{ISNR} \equiv \mathrm{PSNR}(\hbx; \bx) - \mathrm{PSNR}(\tbx; \bx),
$
where $\hbx$ is the image estimated by the proposed method, and $\tbx$ is the image estimated by the method to compare (i.e., (a), (b), (c), and (d) in these experiments). We see that the ISNRs of the images estimated by the proposed method were mostly better than those by the other methods. In the subjective visual comparison in Fig. 4, we also see that the edges are not overemphasized in the images estimated by the proposed method compared to those in the images estimated by other methods.

The calculation times with the proposed method and with methods (b), (c) and (d) for each estimate, using an Intel Core i7 2600 processor, were almost the same, about $20$ minutes.
\begin{figure}[t]
\centering
    {\includegraphics[width=15mm]{InferredKatsukiBest.eps}
	\label{InferredProposed}}
    {\includegraphics[width=15mm]{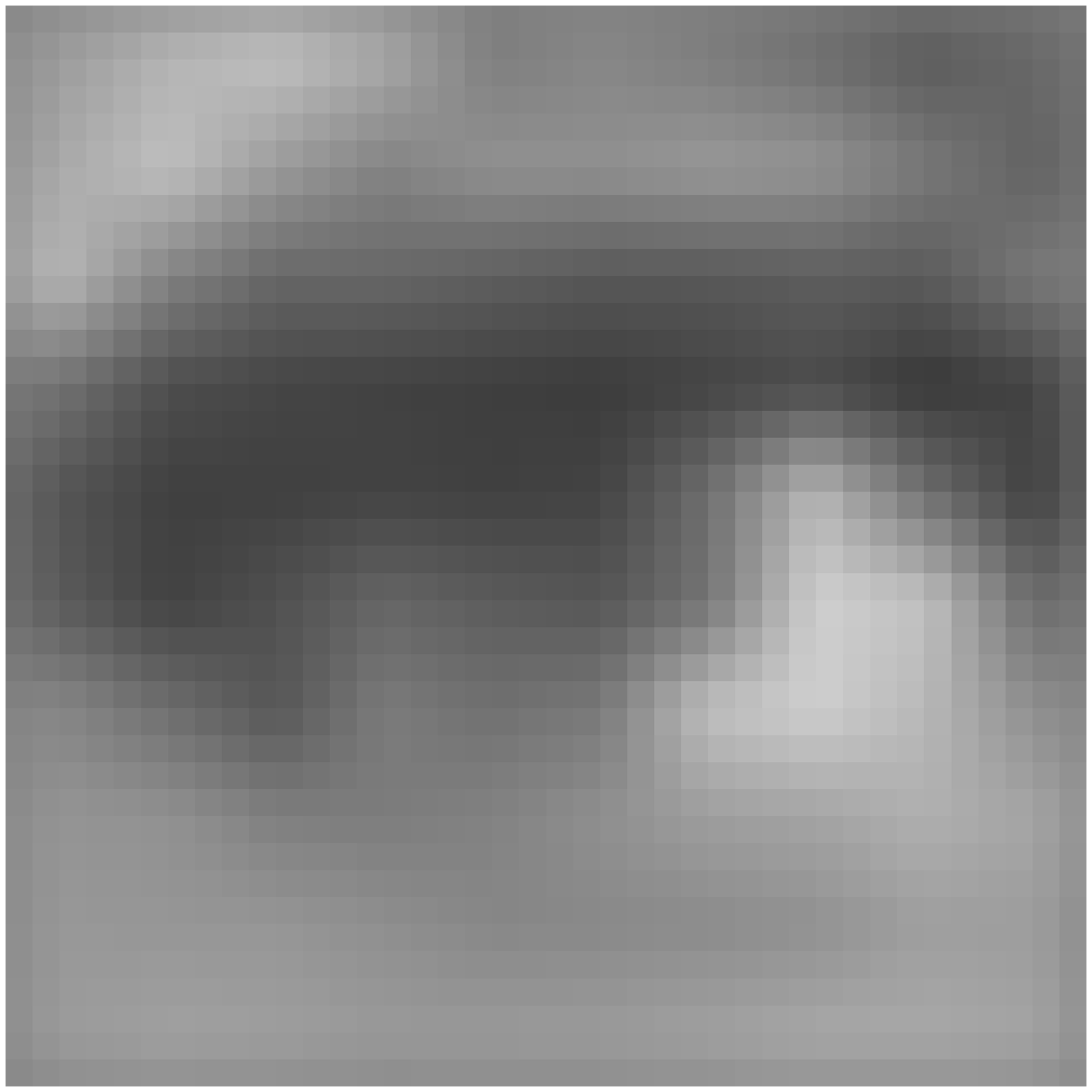}
	\label{InferredKanemura}}
	{\includegraphics[width=15mm]{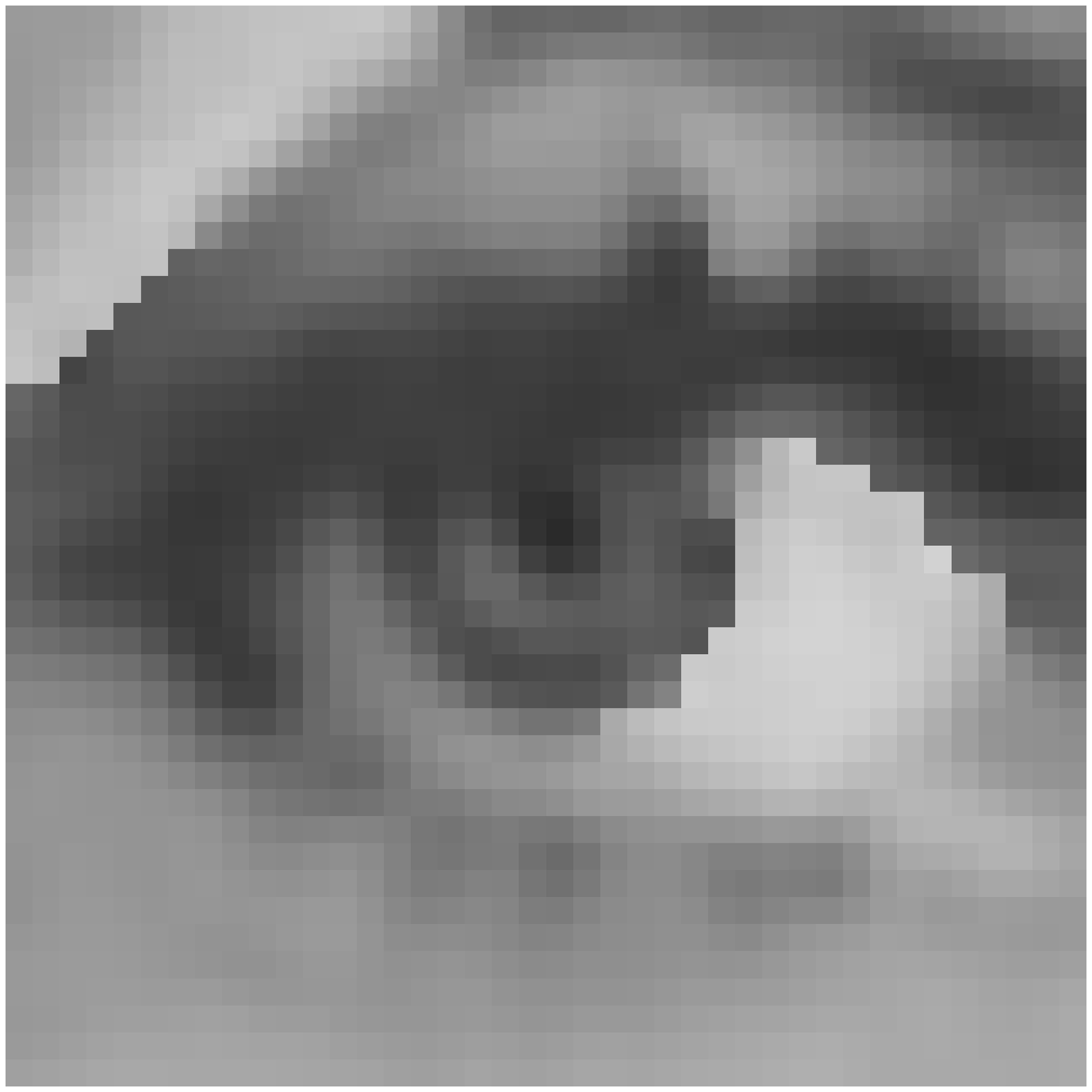}
	\label{InferredKanemura}}
	{\includegraphics[width=15mm]{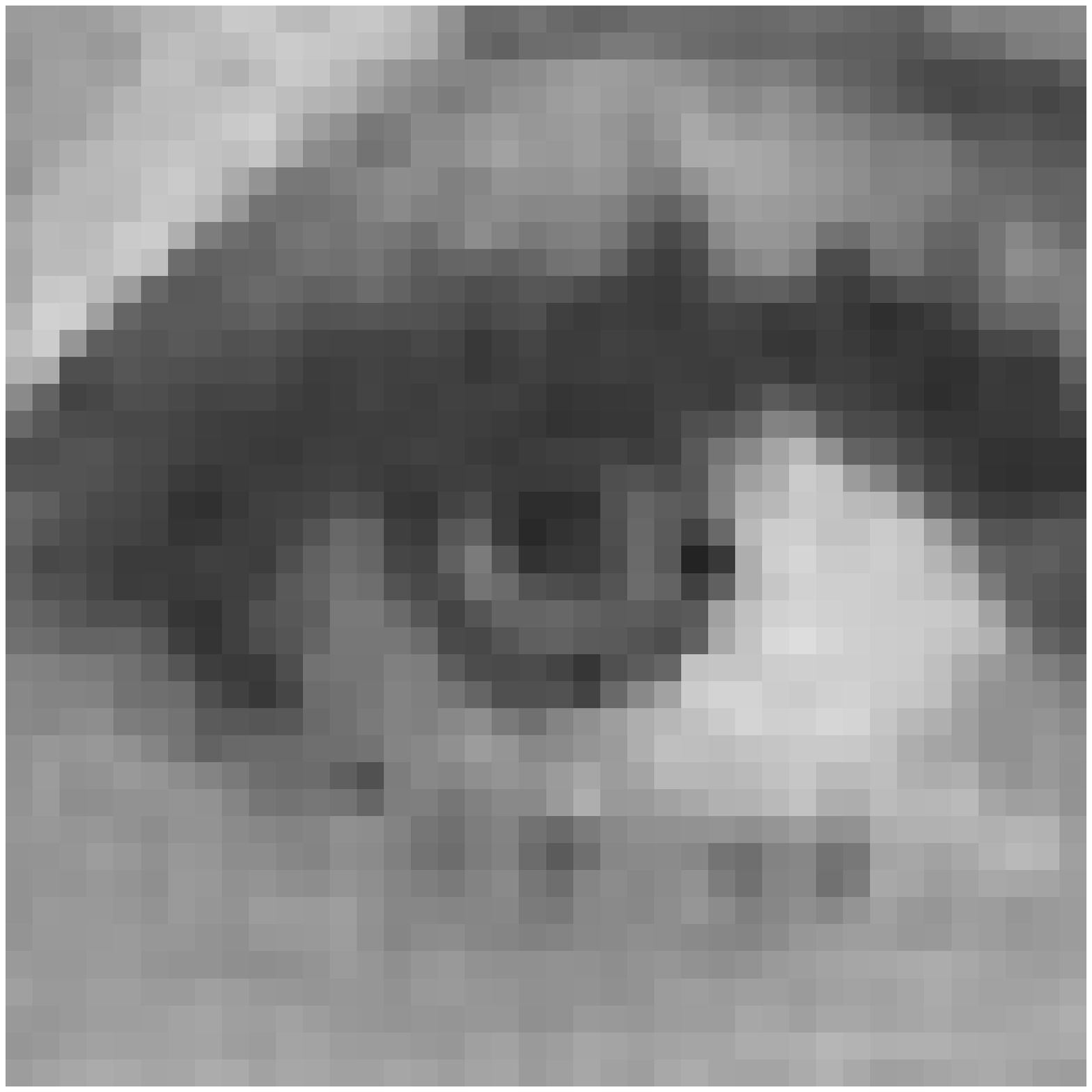}
	\label{InferredMolina}}
	{\includegraphics[width=15mm]{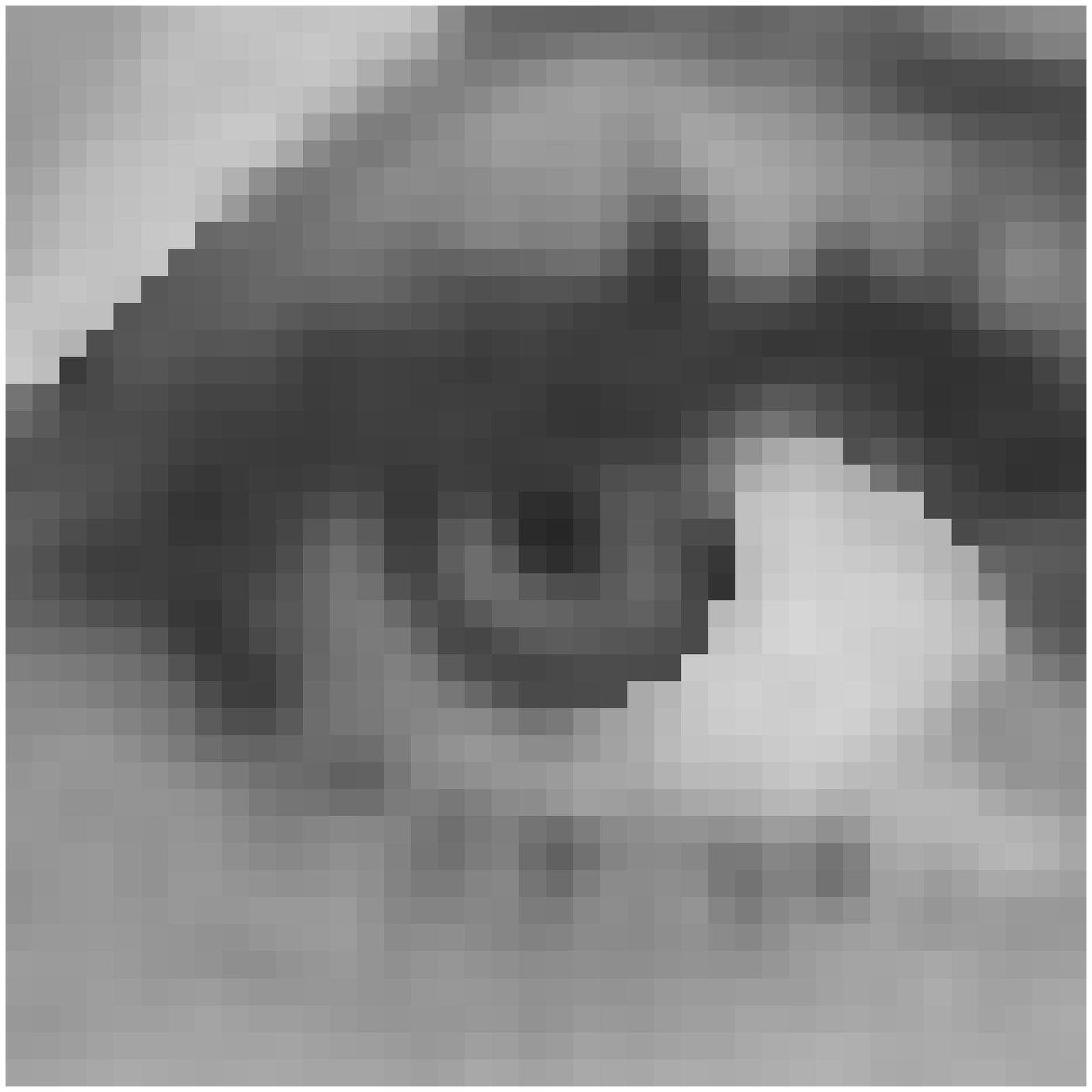}\vspace{-8pt}
	\label{InferredKatsuki}}
\caption{Comparison of estimated images with the proposed method and methods (a), (b), (c), and (d) under SNR$=30$dB}\vspace{-6pt}
	\label{FigInCompared}
\end{figure}

\section{DISCUSSION}
In this manuscript, we have shown how we can adopt a compound Gaussian MRF prior for PM SR. We got good results for almost all images and noise conditions. In the comparison with method (c) and (d), the superiority of ISNR was fair and the inferiority of ISNR was small. Regarding these some unfavorable results in Table \ref{TableResultsPSNR}, we think that the reason is our numerical optimization method falls slightly short of optimization because the proposed method uses more approximations than method (d) and (c). In addition, we consider the result compared to method (c) in the case of the Pepper image in $40$ dB noise is caused by unstable estimation of $\gamma$ and $\rho$, where method (c) fixed the value of $\gamma$ to the true expected value in our implementation same to \cite{Katsuki2011}. Therefore, we think compound Gaussian MRF prior is considered preferable to a ``causal'' Gaussian MRF prior as a natural image prior.

Regarding the estimator, we used the optimal estimator, the PM. From the experimental results, we see that the SR methods with the PM estimator (i.e., the proposed method and method (d)) were more accurate than the SR methods with other estimators (i.e., method (b) and method (c)). This indicates that PM is an optimal estimator.

Regarding the calculation cost, our algorithm requires $\calO(N_\bx^3)$, and we must make our algorithm faster to apply this method to larger images. By using an idea similar to that of \cite{Babacan2011, Katsuki2011}, we have developed a faster algorithm, but this algorithm causes obvious degradation of accuracy. We continue to search for a more effective way to reduce the calculation cost with less degradation of accuracy.

We can say that the proposed method is an SR method with a favorable model and an optimal estimator. In addition, the proposed method does not need any parameter tuning, which is a favorable property in practice. We think our approach to the problem about the conjugate prior and the exponential-order calculation cost can be applied to many other problems, and we will try to do this in our future work.

\end{document}